\documentclass[manuscript,screen]{acmart}
\AtBeginDocument{%
  \providecommand\BibTeX{{%
    \normalfont B\kern-0.5em{\scshape i\kern-0.25em b}\kern-0.8em\TeX}}}

\usepackage{multirow}
\copyrightyear{2022}
\acmYear{2022}
\acmDOI{XXXXXXX.XXXXXXX}

\acmConference[preprint]{preprint}{}{}
\acmBooktitle{Preprint}

\begin{document}

\title{Is Lip Region-of-Interest Sufficient for
Lipreading?}

\author{Jing-Xuan Zhang}
\email{jxzhang27@iflytek.com}
\affiliation{%
  \institution{iFLYTEK Research, iFLYTEK Co. Ltd}
  \city{Hefei}
  \state{Anhui}
  \country{P.R.China}
}
\author{Gen-Shun Wan}
\email{gswan@iflytek.com}
\affiliation{%
  \institution{iFLYTEK Research, iFLYTEK Co. Ltd}
  \city{Hefei}
  \state{Anhui}
  \country{P.R.China}
}

\author{Jia Pan}
\email{jiapan@iflytek.com}
\affiliation{%
  \institution{iFLYTEK Research, iFLYTEK Co. Ltd}
  \city{Hefei}
  \state{Anhui}
  \country{P.R.China}
}








\renewcommand{\shortauthors}{Zhang et al.}

\begin{abstract}
Lip region-of-interest (ROI) is conventionally used for
visual input in the lipreading task.
Few works have adopted the entire face as visual input because
lip-excluded parts of the face
are
usually considered to be redundant and irrelevant to visual speech recognition.
However, faces contain much more detailed information than lips,
such as speakers' head pose, emotion, identity etc.
We argue that such information 
might benefit visual speech recognition if
a powerful feature extractor employing the entire face is trained.
In this work, we propose to adopt the entire face for lipreading with self-supervised learning.
AV-HuBERT, an audio-visual multi-modal self-supervised learning framework,  was adopted in our experiments.
Our experimental results showed that adopting the entire face achieved 16\% relative word error rate (WER) reduction on the lipreading task,  compared with the baseline method using lip as visual input.
Without self-supervised pretraining,
the model with face input achieved a higher WER than that using lip input
in the case of limited training data (30 hours),
while a slightly lower WER when using large amount of training data (433 hours).

\end{abstract}





\maketitle

\section{Introduction}
\label{sec:introduction}

Generation of human speech involves the movement of lip, teeth, jaw, and facial muscles. Therefore face videos of the speaker carry the information about the speech content.
In the machine learning domain, researchers have been working on  representation learning from talking videos, which can be used in many applications~\cite{mroueh2015deep,petridis2018end,petridis2018audio,afouras2018deep,zhou2011towards,wand2016lipreading,luettin1996speaker,rekik2015unified,hao2020survey,Zhang_Richmond_Ling_Dai_2021}. For example, visual speech recognition (VSR), also known as lipreading,  aims at predicting speech content from visual input without relying on audio stream~\cite{zhou2011towards,wand2016lipreading,hao2020survey,noda2014lipreading,martinez2020lipreading}. This can be used for communication in noisy environment or for improved hearing aids. Also, lip videos have been proposed to be used in the security field for biometric authentication~\cite{luettin1996speaker,rekik2015unified,hao2020survey}.

In the lipreading task, lip region-of-interest (ROI) is conventionally adopted as the visual input. For extracting lip ROI, the face detection algorithm is first employed for predicting bounding boxes of the face region. Then facial landmarks are detected from faces and the desired lip regions are cropped. For obtaining visual features,
early works adopted linear transformation on the lip region for dimension-reduction, such as principal component analysis (PCA)~\cite{lee2004design} or discrete cosine transform (DCT)~\cite{sterpu2018towards}. With the rise of deep learning in recent years, it's common to utilize a deep neural network as the feature extractor, which is learned jointly with the classifier by
gradients back-propagation algorithm~\cite{martinez2020lipreading,noda2014lipreading,wand2016lipreading}. Recently, self-supervised learning has emerged as a paradigm which attempts to extract general representations from unlabelled data~\cite{baevski2020wav2vec,hsu2021hubert,bao2022beit}. Shi et al. proposed AV-HuBERT~\cite{shi2022learning}, which learned powerful audio-visual speech representation and achieved state-of-the-art results in the  lipreading task, significantly surpassing supervised trained models.

To our knowledge, few works have adopted the entire face as visual input for lipreading, and lip-excluded parts of the face are usually considered redundant and irrelevant to visual speech content~\cite{Themos2017combine,hao2020survey}.
Vid2Speech~\cite{ephrat2017vid2speech} used the entire face as input for a lip-to-speech reconstruction model. A single speaker was adopted in their experiments thus it was unable to prove the generalization ability to various unseen speakers. Our work is in line with Zhang et al.~\cite{zhang2020can},  in which they comprehensively evaluated the effects of different facial regions for lipreading and showed that the model can benefit from additional clues within extraoral regions. 
However, our work differs from theirs in taking a larger region of face videos into consideration, including
jaw and lower neck.
Also, the discovery hasn't been investigated under the framework of recently proposed self-supervised learning, since their model is constructed based on conventional supervised training.

In conventional supervised training, the entire face may harm the
performance of lipreading by
introducing redundancy. The model may suffer from  the overfitting effect and be difficult to generalize on new speakers, especially when training data is insufficient.
This paper proposed to solve this issue by adopting self-supervised pretraining.
Without the requirement of transcriptions, the pretraining data is easier to  be collected.
Self-supervised learning forces the model to capture long-range dependency and the internal data structure effectively by solving the mask prediction~\cite{hsu2021hubert} or contrastive prediction~\cite{baevski2020wav2vec} task. Hence we expect that lipreading can potentially benefit from the rich details of the entire face, and avoid the overfitting effect in the meantime.
Another advantage of using the face input is that the landmark detection and lip cropping process can be
removed, in both data preparation and testing stage. This simplifies the training pipeline and saves the computational cost.

In our experiments, AV-HuBERT was adopted as self-supervised training framework.
The method using the entire face and that using lip ROI as visual input for AV-HuBERT were compared. A large-scale audio-visual dataset, LRS3~\cite{Afouras18d},  is used in this study, which is very challenging since its large vocabulary size ($\sim$4M) and in-the-wild face videos with large variations in head pose.
Our experimental results showed that the method with face input achieved significantly better performance than that with lip ROI input in the lipreading task, reducing WER by 16\% relatively.
The speaker identity information contained in the entire face
can be one of the reasons for better lipreading results, and
the speaker verification task was performed for measuring such information.
We further compared models with face input and that with lip input without self-supervised pretraining. Despite the model with face input achieved higher WER with
limited training data (30 hours) than that using lip, the gap narrowed as the amount of
data increased. Eventually, the model with face input obtained slightly better performance
than that with lip ROI input (433 hours).

\section{AV-HuBERT}
\label{sec:av_hubert}

\begin{figure}[h]
    \centering
    \includegraphics[width=0.7\linewidth]{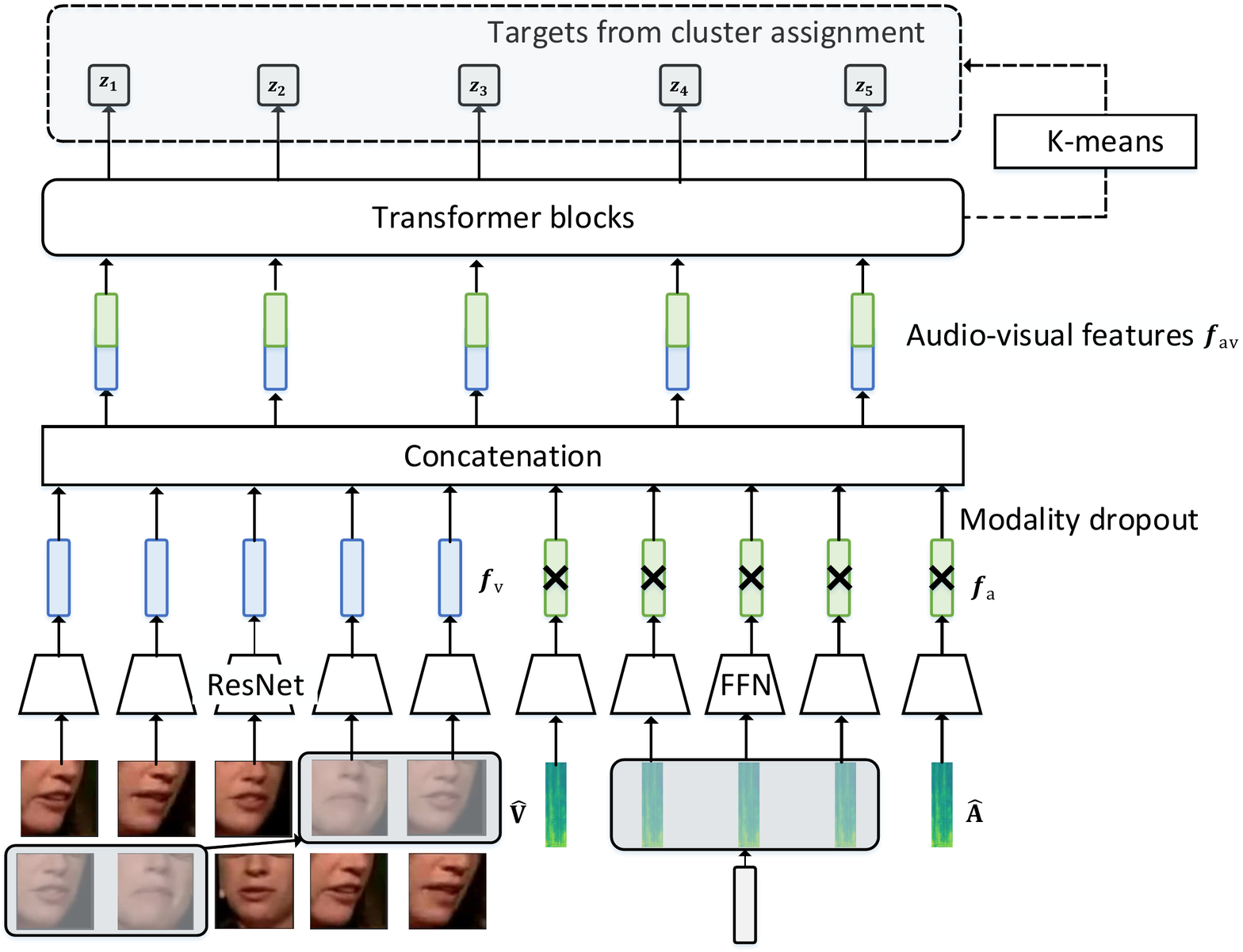}
    \caption{Overview of the AV-HuBERT model~\cite{shi2022learning}.}
    \Description{Overview of the AV-HuBERT model structure.}
    \label{fig:avhubert}
\end{figure}

In this section, we review the method of audio-visual HuBERT (AV-HuBERT)~\cite{shi2022learning}. AV-HuBERT is an
extended version of HuBERT~\cite{hsu2021hubert} from learning speech representation to
learning audio-visual representation jointly. It consumes
both acoustic and video frames for the mask prediction training, which enables  modeling and distillation of the correlation between the two modalities.
For obtaining targets of mask prediction, the multi-modal hidden units are automatically discovered and iteratively refined. The model is illustrated in Figure~\ref{fig:avhubert}.

For audio frames $\textbf{A} = [\textbf{a}_1, \textbf{a}_2, \dots, \textbf{a}_T]$, an audio mask $M_a$ is sampled randomly. Audio frames are masked by replacing the audio frame $\textbf{a}_i$ with a learned embedding $\textbf{e}_a$ for each $i \in M_a$, yielding the corrupted audio $\hat{\textbf{A}}$. For video
frames $\textbf{V} = [\textbf{v}_1,\textbf{v}_2,\dots, \textbf{v}_T]$, video mask $M_v$ is generated and
video frames in
each masking
span are substituted with a random segment of the same video. The corrupted video is denoted by $\hat{\textbf{V}}$. Audio and video streams are masked independently with a probability $p_{maska}$ and $p_{maskv}$ respectively.  The corrupted audio and video are passed through a linear projection layer and a ResNet to obtain intermediate features $\textbf{f}_a$ and $\textbf{f}_v$ respectively. A modality dropout is then employed to
train the model in absence of audio or video stream.  With a probability $p_m$, both modalities are used. The probability of selecting audio is $p_a$ when only one modality is used, and features of the other modality are set to be all zeros.
$\textbf{f}_a$ and $\textbf{f}_v$ are channel-wise concatenated to form the
audio-visual representation $\textbf{f}_{av}$.
Then it is fed into a transformer encoder for generating multi-modal contextualized features,
which are used for predicting pseudo labels derived from cluster assignment. The clusters are generated from MFCC or hidden outputs of the transformer encoder in the initial and following iterations respectively. For unsupervised clustering, k-means algorithm is used~\cite{arthur2006k}. The pretraining loss of AV-HuBERT is
\begin{equation}
L = - \sum_{t \in M_a \cup M_v }{
\text{log}(p_t(z_t))
} - \alpha \sum_{t \notin M_a \cup M_v }{
\text{log}(p_t(z_t))
},
\label{eq:eq2}
\end{equation}
where $\textbf{p}_{1:T}$ are output probabilities and $\textbf{z}_{1:T}$ are
pseudo labels. $\alpha$ controls the contribution of unmasked region
in overall objective. The masked prediction loss forces model to learn high-level representation of unmasked inputs to infer the targets of masked ones correctly. The model iterates over the feature clustering and masked prediction steps until the model's performance no longer improves.
After pretraining, the cluster prediction head is removed and
the output features from the transformer encoder can be used for downstream tasks, such as lipreading.

\section{Methodology}

\begin{figure}[t]
\centering
\includegraphics[width=0.9\linewidth]{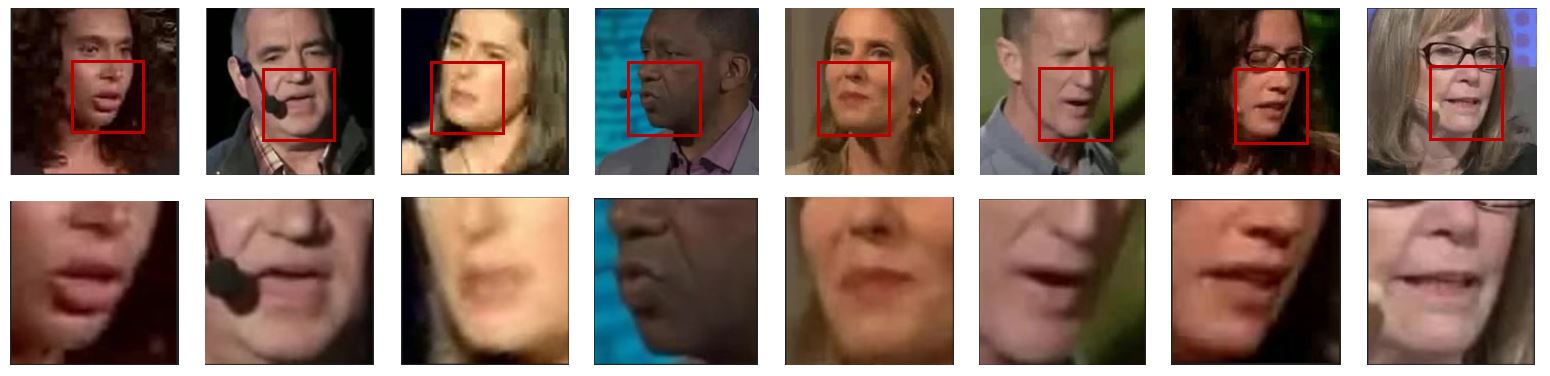} 
\caption{Example images of the face and the extracted lip ROI.}
\Description{Example images of the face and the extracted lip.}
\label{fig:face_lip}
\end{figure}

For the baseline method, lip region-of-interest (ROI) is extracted.
68 facial keypoints are detected and aligned to a reference face via affine
transformation. 96 $\times$ 96-pixel region-of-interest is cropped centered on the mouth from the original 224 $\times$ 224-pixel image of the entire face. Examples of the original face
and the extracted lip ROI are presented in
Figure~\ref{fig:face_lip}. The cropped patches are then processed with data augmentation, including random cropping and horizontal flipping. They're further converted to grey-scale and normalized to obtain the final inputs of the model. For our proposed method,
images of the entire face are used. Face images are resized into 96 $\times$ 96-pixel, then
processed by data augmentation, grey-scale conversion, and normalization. For
the audio stream, log filterbank features are used.

Notice that adopting face rather lip ROI not only changes the inputs of visual stream, but also
affects the targets during pretraining. For our proposed method, audio-visual hidden features are derived from the entire face and audio in the feature clustering step, in contrast with the lip ROI and audio for the baseline. It's expected that faces not only provide richer details than lips in masked prediction, but also in favor of producing higher quality cluster assignments at the feature clustering stage.

After pretraining, the model is finetuned on the lipreading task. A sequence-to-sequence
structure is adopted, and  the pretrained model is used as the encoder. The transformer decoder is  randomly initialized and added on the top of the encoder. Parameters of the pretrained encoder
are frozen for the initial $k$ steps and then jointly optimized with the decoder for the
rest steps of finetuning.


\section{Experiments}

\subsection{Implementation details}

Experiments were conducted on the LRS3 dataset~\cite{Afouras18d}, which contains around 433 hours of English videos.
LRS3 consists of pretraining, train-validation, and test set
with 118516/31982/1321 utterances, and 5090/4004/412 speakers respectively.
We preprocessed the pretraining set by splitting long utterances into shorter clips with a
limitation of 15 seconds maximum duration. Then pretraining and train-validation sets
were merged, from which 1200 utterances were kept for validation and the rest were for training.
The AV-HuBERT model was pretrained on all the training data. A subset
of or the whole training data was used for finetuning.

The image encoder was based on the modified ResNet-18 structure \cite{afouras2018deep,ma2021end}, and the audio encoder was a linear projection layer. The ``BASE'' AV-HuBERT structure~\cite{shi2022learning} was used which had 12 transformer blocks.
$\alpha$ in Equation~\ref{eq:eq2} was set to 0. $p_m$ and $p_a$ were set to 0.5. Masking percentage
for video and audio stream were $80\%$ and $30\%$ respectively. The number of transformer blocks
was 6 for the decoder in lipreading.
To measure the speaker information contained in the
visual representation, speaker verification models
were also constructed in our experiments.
Visual representation from the 4th transformer block
of the pretrained model
 was passed through a $3\times512$ DNN for speaker
classification, and the pretrained model was frozen throughout finetuning process.
During testing, hidden activations before the output layer were extracted and then $L2$ normalized as the speaker embedding~\cite{variani2014deep}. Cosine distance was used for
comparing two speaker embedding vectors.


AV-HuBERT performs a total of 5 iterations in the original paper. To save the training time and computation cost, our experiments started from the 4th iteration and the pretrained model checkpoint publicly released by the author
was used\footnote{\url{https://facebookresearch.github.io/av_hubert/}}.
Our experiments were implemented with fairseq~\cite{ott2019fairseq}.
We conducted the 5th iteration for the baseline and our experiments showed no further improvement
by another iteration. For our proposed method,
we conducted the 5th iteration with face input, while the targets from cluster assignment were the same as the baseline.
After the 5th iteration, faces rather than the lip can be used as visual input in feature clustering
stage for generating targets.
Then another iteration proceeded with face-derived targets to get the best performance.
The number of k-means clusters was 2000 for both the 5th and 6th iteration.
For other training details, we followed the setting of AV-HuBERT~\cite{shi2022learning}.

\subsection{Lipreading with pretraining}
\label{subsec:lipreadingwithpretraining}

\begin{table}
     \centering
     \caption{Word Error Rate (WER) of lipreading on the LRS3 test set.
     ``Input'' and ``Target'' represent
     using face or lip as model's visual input
     and  for pretraining's targets generation respectively.
     $^\dagger$Our reproduction results are better than the reported WER of
     51.7\% and 44.0\% for
     30 and 433 hours of labelled data respectively in the original paper~\cite{shi2022learning} .}
    \label{tab:tab1}
    \begin{tabular}{c c c c  c}
    \toprule
    \textbf{Method} &  \textbf{Labelled data} (hrs) &
       \textbf{Input} &
       \textbf{Target} &
     \textbf{WER} (\%) \\
    \midrule
    Afouras et al.~\cite{afouras2020asr} & 590 & lip & - & 68.8 \\
    Afouras et al.~\cite{afouras2018deep} & 1519 & lip & - & 58.9 \\
    Xu et al.~\cite{xu2020discriminative} & 590 & lip & - & 57.9 \\
    Ma et al.~\cite{ma2021end} & 433 & lip & - & 46.9 \\
    Ma et al.~\cite{ma2021end} & 590 & lip & - & 43.3 \\
    Makino et al.~\cite{makino2019recurrent} & 31K & lip & - & 33.6 \\
    \hline
    \multirow{6}{*}{AV-HuBERT} &
    \multirow{3}{*}{30} & lip & lip & $^\dagger$47.7 \\
    & & \emph{face} & lip & 42.5 \\
    & & \emph{face} & \emph{face} & \textbf{39.9} \\
    \cline{2-5}
    &   \multirow{3}{*}{433} & lip & lip &  $^\dagger$40.3 \\
    & & \emph{face} & lip & 34.6 \\
    & & \emph{face} & \emph{face} & \textbf{33.8} \\
    \bottomrule
    \end{tabular}
\end{table}

\begin{table}[t]
     \caption{Equal Error Rate (EER) (\%) results of speaker verification on the LRS3 test set.
    }\label{tab:sv}
    \centering
    \begin{tabular}{c  c  c}
    \toprule
     \multirow{2}{*}{\textbf{Input}}  &  \multicolumn{2}{c}{ \textbf{Labelled data} (hrs)} \\
     \cline {2-3}
     &  30 &  433 \\
    \midrule
    lip &  2.03 &   1.03\\
    \emph{face} & \textbf{1.70} & \textbf{0.76} \\
    \bottomrule
    \end{tabular}
\end{table}

\begin{table}[t]
     \caption{WER (\%) results of ablation study on different regions of the face.
    }\label{tab:ablation}
    \centering
    \begin{tabular}{c  c  c  c}
    \toprule
      \textbf{face} & \textbf{face-eye} & \textbf{face-eye-neck} & \textbf{face-eye-neck-side} \\
    \midrule
    42.5 & 42.7 &  44.1 &  47.6 \\
    \bottomrule
    \end{tabular}
\end{table}

This section compares models using the entire face and those using the lip ROI in the lipreading task based on self-supervised pretraining.
Pretrained models were finetuned on 30 and 433 hours of training data respectively.
Our experimental results and the reported results of previous works are presented in Table~\ref{tab:tab1}.
For the proposed method, we first replaced the lip ROI with face for visual input in the 5th iteration, while the targets were still derived from the lip ROI and audio (``\textbf{Input} \emph{face}, \textbf{Target} lip''). Then the targets
were refined by using face as visual input in features clustering. Therefore,
both input and target adopted face for the 6th iteration (``\textbf{Input} \emph{face}, \textbf{Target} \emph{face}'').

From Table~\ref{tab:tab1}, AV-HuBERT pretraining effectively improved
the performance of lipreading task and achieved lower WER compared with
previous works using the same amount of data.
For methods based on AV-HuBERT pretraining,
we observed that solely using face as visual input achieved significantly
better results than the lip ROI.
When the targets were further refined using face input, WER was further reduced.
The experiment results indicated that using face for both model inputs and targets generation achieves the best performance. Eventually,
our proposed method achieved 16\% relative WER reduction compared with the baseline using lip ROI, under both configurations of 30 and 433 hours labelled data.
Therefore, it demonstrates the pretrained model can effectively exploit the entire face
input and achieves better lipreading performance than that using traditional lip ROI.

Researches have shown that
the ASR model's performance can be improved by speaker information~\cite{karafiat2011ivector,rouvier2014speaker,sari2020unsupervised}.
Similarly, more speaker information provided by face can be one of the factors leading to superior lipreading results.
To measure the speaker identity information contained in visual representation, speaker verification models
were constructed.
Equal error rate (EER) results of our experiments are summarized in Table~\ref{tab:sv}.
 The method using face as visual input
 achieved lower EER than the equivalent one using lip ROI,
 indicating  more speaker-specific information is contained
 in the face.

 Ablation study was further conducted for investigating the importance of
 different regions of the face. The
 top 30\% (``\textbf{-eye}''), bottom 25\%  (``\textbf{-neck}''), as well as 27.5\% left and right sides (``\textbf{-side}'') of
 the face images were gradually masked during both pretraining and finetuning. The experiments were
 based on the 5th iteration of pretraining with 30-hour labelled data for finetuning, and the results are presented in Table~\ref{tab:ablation}. From the table, the neck and two side areas of the face image contribute the most for lipreading except the lip ROI.

\subsection{Lipreading without pretraining}

\begin{figure}[t]
\centering\includegraphics[width=0.5\linewidth]{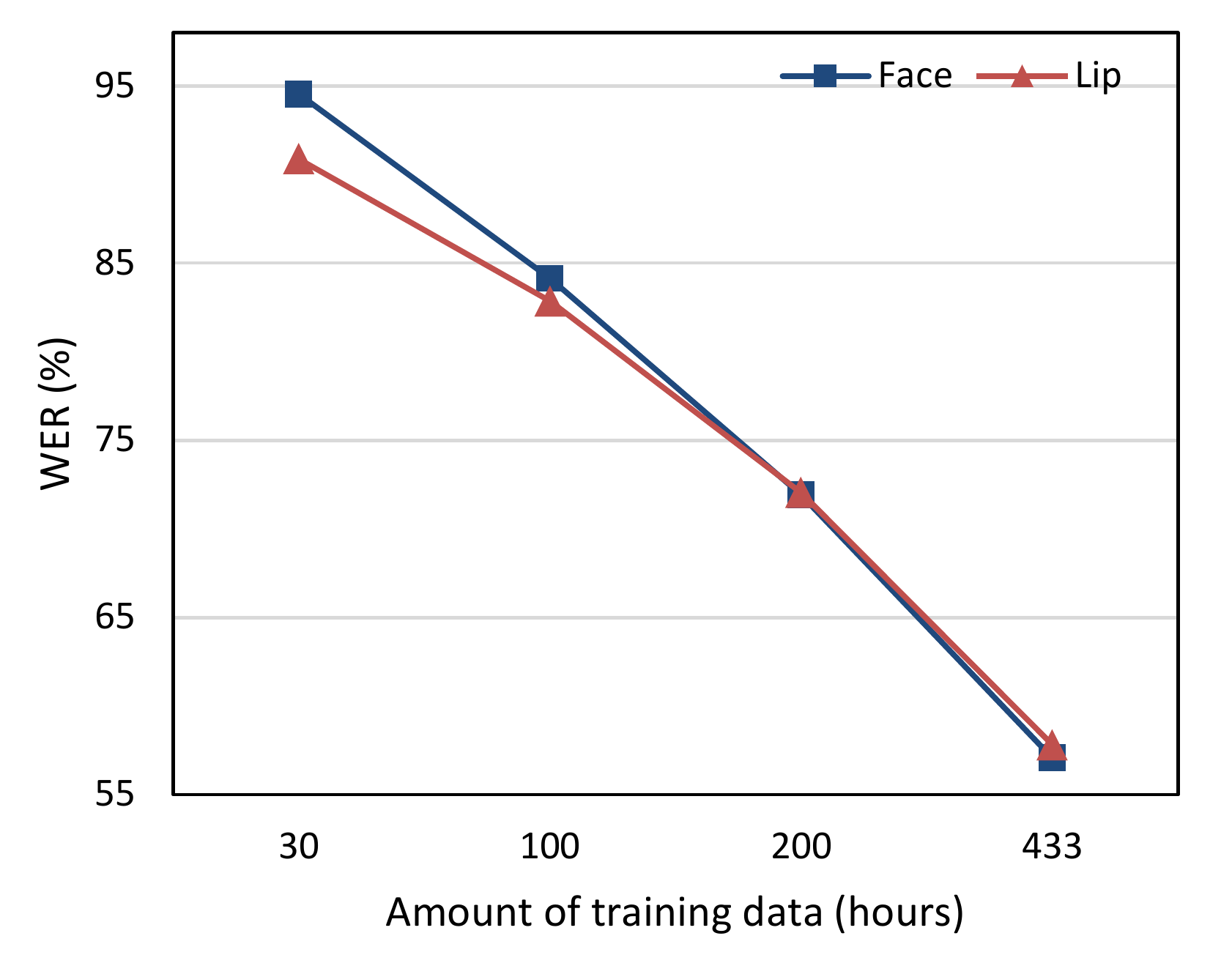}
\caption{WER results of lipreading given different amounts of training data without
pretraining.}
\Description{Word error rate curve which compares the face based model and the lip based model.}
\label{fig:wer}
\end{figure}

\begin{figure*}[t]
\centering
\includegraphics[width=1.0\linewidth]{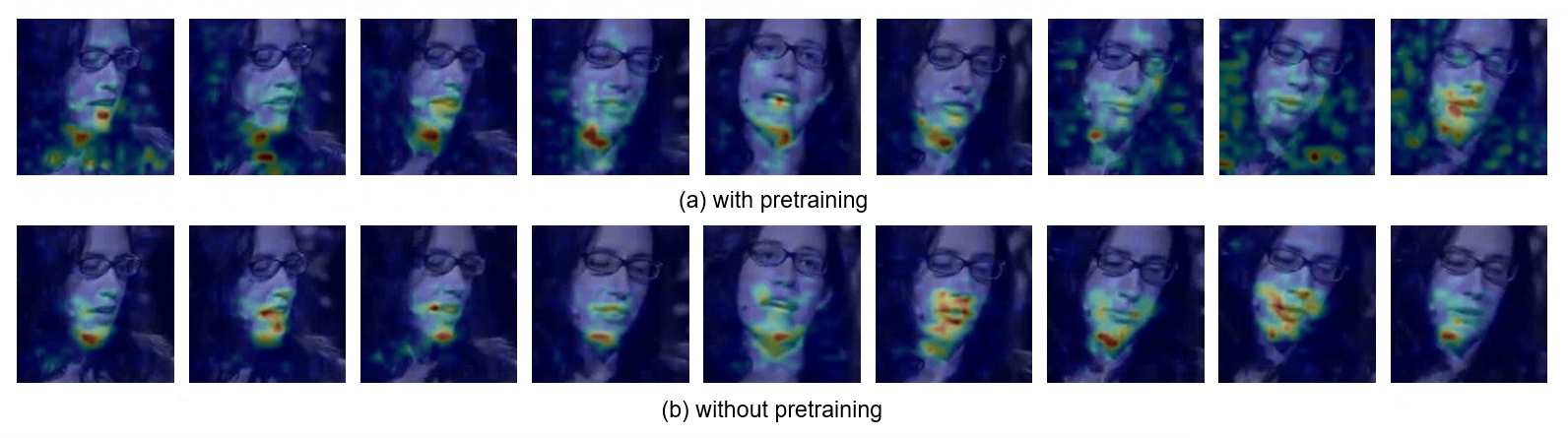} 
\caption{Gradients visualization of the model (a) with self-supervised pretraining (b) without pretraining. Face images are shown in grey scale and the
colored gradient maps overlay on the face images.}
\Description{Gradients visualization of the models. Face images are shown in grey scale and the colored gradient maps overlay on the face images.}
\label{fig:grad}
\end{figure*}

Next, whether the model with face input outperforms the lip one without pretraining was investigated.
The encoder parameters were randomly initialized and jointly trained
with the decoder from the beginning. Experiments were conducted on different amounts of training data, including 30, 100, 200 and 433 hours, and the results are summarized in Figure~\ref{fig:wer}.
From this figure, the model using face as visual input achieved a higher WER than lip ROI when only 30-hour training data available. However, the gap narrowed as we increased the amount of
training data. The model using face input eventually achieved slightly lower WER than that using lip
with 433h training data (57.1\% vs 57.8\%). 

Face as visual input introduces redundancy and brings challenges of various head poses and speaker-dependency for training the model. Therefore the model tends to be overfitting with limited data available for training.
More training data alleviates this effect and enables the model to capture correlations between
lip-excluded parts of face and the speech content.
Nevertheless, even all training data was used, the model using face for lipreading without pretraining did not
significantly outperform that using lip as in Section~\ref{subsec:lipreadingwithpretraining}. This indicates the advantage of self-supervised pretraining
for learning a powerful feature extractor. We further visualized gradients of face images and compared the model with
pretraining to that without, and an example are shown in  Figure~\ref{fig:grad}. It's observed that gradients concentrated on the mouth region
for the randomly initialized model. 
While gradients spread to face and neck regions other than the lip
for the pretrained model. This implies the pretrained model can effectively make use of
the entire face rather than only focus on the lip ROI.



\section{Conclusion}

In this paper, we propose to adopt the entire face for visual speech
representation learning based on self-supervised learning framework.
Compared to lip ROI, the entire face contains richer information such as
the speaker's head pose, emotion, identity, and so on. Such information might benefit
lipreading task but also introduce redundancy that may lead to the overfitting effect.
With self-supervised pretraining,
the model is able to exploit face videos efficiently for visual speech recognition.
And this was proved by our experiments which showed the model with face input achieved significantly better
performance than that with traditional lip ROI input.
In the future work, we will explore  incorporating a wider variety of visual cues such as body movements for lipreading.

\bibliographystyle{ACM-Reference-Format}
\bibliography{sample-base}


\end{document}